\documentclass[letterpaper]{article} 
\usepackage[preprint]{aaai2027}
\usepackage[hyphens]{url}  
\usepackage{graphicx} 
\usepackage{natbib}  
\usepackage{caption} 
\usepackage{booktabs}

\usepackage{amsmath}
\usepackage{amssymb}

\title{PACE: Phase-Progress-Aware Credit for Long-Horizon Embodied Manipulation}
\author{
    Chengye Song\textsuperscript{\rm 1}\equalcontrib,
    Jiawei Zhang\textsuperscript{\rm 2}\equalcontrib,
    Rui Song\textsuperscript{\rm 1},
    Shengqi Wang\textsuperscript{\rm 1},\\
    Xiangrong Zhang\textsuperscript{\rm 2},
    Ziyi Wang\textsuperscript{\rm 1},
    Huanbin Zhou\textsuperscript{\rm 2},
    Hongzhou Wang\textsuperscript{\rm 2}\corresponding
}
\affiliations{
    \textsuperscript{\rm 1}Intelligence Science and Technology, Dalian University of Technology,\\
    No. 2 Linggong Road, Ganjingzi District, Dalian 116024, China\\
    \textsuperscript{\rm 2}Jilin University, No. 2699 Qianjin Street, Changchun 130012, China\\
    wanghongzhou@jlu.edu.cn
}

\begin{document}

\maketitle

\begin{abstract}
Post-training of vision-language-action (VLA) models typically relies on expert demonstrations and policy interaction trajectories. However, in long-horizon manipulation, a single episode often spans hundreds of control steps and multiple phases, while success or failure is only revealed at episode termination. Policy improvement therefore requires step-level credit signals to distinguish behaviors that advance the task from those that stall or regress. We present PACE, a credit-assignment framework for post-training on long-horizon manipulation, centered on a phase-progress-aware critic. PACE consists of two key modules: (1) the Global-Local Cooperative Value-Correction Critic (GLC-Critic) aggregates visual and motion-difference features within local temporal windows to infer the phase and intra-phase progress of each step, and applies residual correction to a discretized remaining-cost distribution accordingly, enabling step-level credit assignment; (2) Progressive Policy Distillation (PPD) converts credit into positive and negative conditions via task-wise thresholds and trains a credit-conditioned action generation policy: it first protects the pretrained policy with high-credit positive samples, then incorporates all positive and negative credits to learn the quality boundary, and at inference amplifies high-credit behaviors through the difference between conditional outputs. Extensive simulation experiments and diverse real-world robotic-arm experiments demonstrate that PACE consistently achieves significant improvements over the strongest baseline.
\end{abstract}


\section{Introduction}

Vision--Language--Action (VLA) models pretrained on large-scale heterogeneous robot data are typically post-trained with expert demonstrations and policy interaction trajectories. In long-horizon manipulation, however, a single episode often spans hundreds of control steps and multiple task phases, while success or failure is only revealed at episode termination. The terminal outcome alone cannot indicate which behaviors advanced the task, which stalled progress, or which undid progress that had already been achieved. Effective post-training therefore requires reliable, fine-grained credit assignment.
\vspace{0pt plus 1fill}

\begin{figure}[!t]
\centering
\includegraphics[width=\columnwidth]{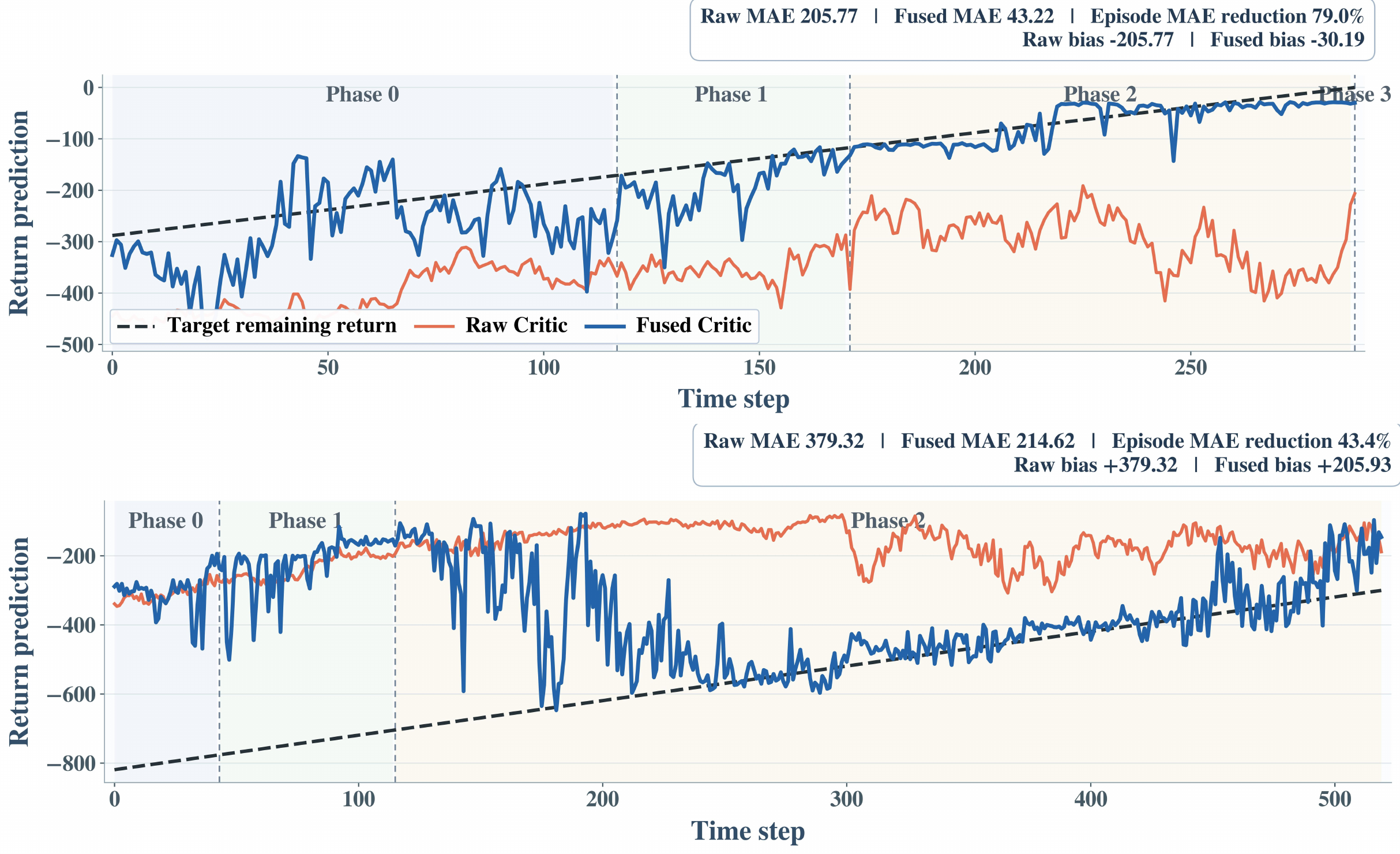}
\caption{Raw and fused remaining-cost predictions on successful (top) and incomplete (bottom) trajectories.}
\label{fig:credit_motivation}
\end{figure}

Recent critic-based robot post-training methods estimate values or relative advantages and use them for offline policy improvement or distillation \citep{amin2025pi06star,mao2026arm,zhai2025vlacritic}. Obtaining a reliable value signal, however, remains difficult for three reasons. First, long-horizon tasks usually provide only sparse terminal rewards, making it hard to determine each timestep's specific contribution to the final outcome. Second, different task phases may share highly similar visual observations; a critic relying on single-frame information struggles to identify phase transitions and changes in contact state, and is prone to unstable value estimates. Finally, value-estimation errors propagate into advantage computation, producing misleading signals that can degrade the downstream policy. Figure~\ref{fig:credit_motivation} illustrates these phase-dependent errors on both successful and incomplete trajectories.

Credit estimation and utilization are also coupled. Early interaction data contain few successful behaviors, so aggressively learning from negative credit can suppress rare but useful actions and destabilize the pretrained policy. As rollout coverage improves, however, negative examples become informative for identifying the boundary between task-advancing and regressive behavior. Long-horizon post-training therefore requires not only phase-aware credit estimation, but also a curriculum that adapts how the resulting credit is used.

PACE addresses long-horizon manipulation with a distributional remaining-cost critic whose expectation is corrected using phase--progress coordinates inferred from short-term dynamics. Phase boundaries come from simulator semantic predicates in simulation and manual annotation on the real robot. Intra-phase progress follows normalized elapsed time except in the final incomplete phase of failed trajectories, where Qwen3-VL estimates are manually reviewed. The state-dependent scalar residual targets phase transitions and contact-state changes without reshaping the base distribution. Progressive Policy Distillation (PPD) uses high-confidence positive-credit samples in early training and incorporates both positive and negative credit later, protecting pretrained behavior before learning the quality boundary. Iterative data collection allows the critic and policy to improve together. The main contributions are listed below.
\begin{itemize}
    \item PACE is a reinforcement-learning post-training framework for long-horizon, multi-phase manipulation that produces step-level credit signals for policy optimization.
    \item The framework combines a distributional remaining-cost critic with phase--progress-aware scalar correction and progressive policy distillation, which adapts the use of positive and negative credit signals as rollout coverage improves.
    \item Evaluation on LIBERO-Long and a limited-scale real dual-arm dataset shows that PACE improves over ReCAP from 73.8\% to 83.3\% in simulation and from 66.5\% to 81.8\% on the real robot while reducing task execution cost.
\end{itemize}

\begin{figure*}[t]
\centering
\makebox[\textwidth][l]{\hspace*{0.08\textwidth}\includegraphics[width=0.92\textwidth]{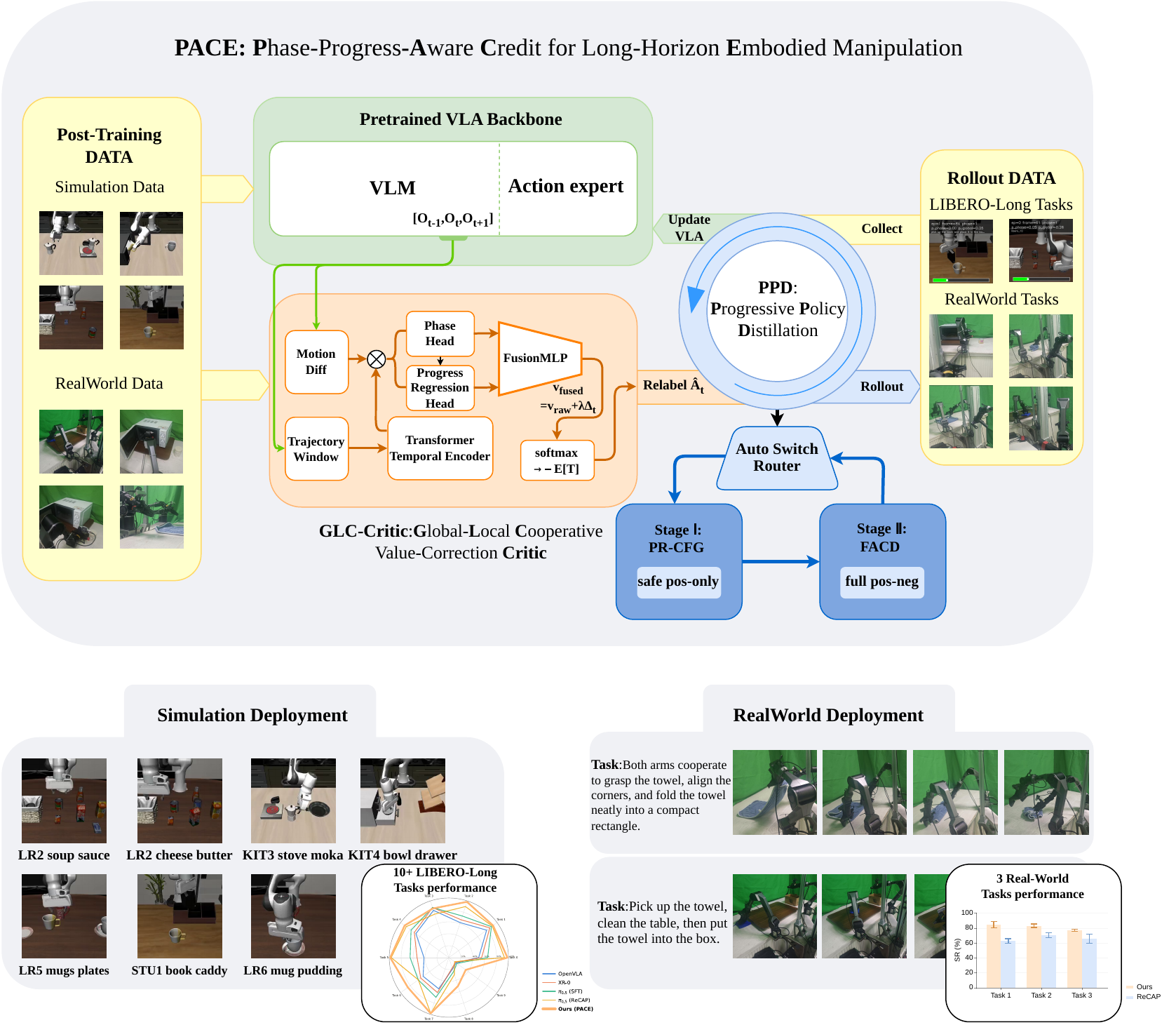}}
\caption{Overview of PACE. Post-training trajectories supervise the phase--progress-aware GLC-Critic, which produces corrected remaining-cost estimates for credit relabeling. Progressive Policy Distillation then adapts the use of positive and negative credit as iterative data collection improves critic coverage, updating the VLA policy for simulation and real-robot deployment.}
\label{fig:pace_overview}
\end{figure*}

\section{Related Work}

\noindent\textbf{Vision-Language-Action Models.}
Recent VLA models, including RT-1, RT-2, OpenVLA, $\pi_0$, $\pi_{0.5}$, and Xiaomi-Robotics-0 \citep{brohan2023rt1,brohan2023rt2,kim2024openvla,black2025pi0,physicalintelligence2025pi05,cai2026xiaomirobotics0}, have evolved mainly along action-generation mechanisms---from autoregressive prediction of discretized action tokens to flow-matching-based continuous action generation---yielding increasingly strong policy backbones for diverse manipulation tasks. Credit assignment, however, is largely orthogonal to these advances: backbone models seldom estimate which steps of a long-horizon trajectory advance, stall, or reverse task progress. Our work builds on such backbones and targets this complementary post-training problem.

\noindent\textbf{Reinforcement Learning and Credit-Conditioned Post-Training.}
Recent works incorporate reinforcement learning into VLA post-training, either through sample-efficient offline or interaction-guided fine-tuning \citep{huang2025corft,su2026igrft,chen2025conrft,dong2026expoft}, or by scaling online RL on simulated manipulation suites \citep{lu2025vlarl,zang2025rlinfvla,chen2025tgrpo}. Experience-based approaches further use advantage-conditioned policies or learned relative-advantage rewards to improve over collected behavior, as in ReCAP and ARM \citep{amin2025pi06star,mao2026arm}; vision-language representations have also been used for value inference and process-level robotic critic modeling \citep{ma2025incontextvalue,zhai2025vlacritic}. These methods are sensitive to credit quality, which is inherently difficult to obtain under long-horizon sparse rewards. In parallel, credit-conditioned policy updates guide imitation or generative policies with trajectory-quality labels or optimality conditions, including classifier-free guidance \citep{ho2022classifierfree}; diffusion guidance itself can be interpreted as a controllable policy-improvement operator \citep{frans2025diffusionguidance}. Applying positive and negative conditions symmetrically from the outset can be risky under scarce data, because imitation-learning performance depends strongly on the quality and composition of the training data \citep{belkhale2023dataquality}. Low-quality negative samples may therefore pull the action distribution away from valid pretrained behaviors.

\noindent\textbf{Temporal Progress Modeling for Long-Horizon Manipulation.}
Long-horizon manipulation naturally calls for temporal abstraction, as classically studied in hierarchical RL \citep{sutton1999between}; adapting such abstractions to the pre-train-then-post-train paradigm of VLAs, however, remains nontrivial. Recent works use stage structure or frame-wise temporal distance to learn denser robotic rewards \citep{chen2026sarm,liu2026timerewarder}, while others pursue progress-aware skill learning and hierarchical imitation \citep{kim2026progvla,buamanee2026bihil}. These methods model temporal progress primarily as a reward, policy-conditioning signal, or decomposition mechanism. In contrast, PACE uses phase--progress coordinates and local dynamic evidence to correct the expected remaining cost of a distributional base critic, then adapts the use of the resulting positive and negative credit during policy distillation.

\section{Method}

PACE converts terminal outcomes into remaining-cost targets, corrects a distributional critic's expected cost using phase--progress-aware local dynamics, and distills the resulting credit through a coverage-adaptive policy curriculum.

\subsection{Remaining-Cost Modeling with Survival Cost}

We formulate each task as an MDP with survival cost $c_t=1$. Successful termination has zero terminal cost, whereas failure incurs $C_{\mathrm{fail}}$:

\begin{equation}
\label{eq:terminal_cost}
c_{\mathrm{term}}(\rho(\tau)) = \begin{cases}
    0, & \rho(\tau)=\text{success}, \\
    C_{\mathrm{fail}}, & \rho(\tau)=\text{failure},
\end{cases}
\end{equation}

Simulation outcomes come from LIBERO task-success predicates, and real-robot outcomes are judged by human evaluators; Qwen3-VL never determines terminal success or failure.

Let $T_\tau$ be the terminal step and $c_t=1$ the per-step cost; the remaining-cost target at time $t$ is

\begin{equation}
\label{eq:remaining_cost_target}
y_t^{(\tau)} = \sum_{s=t}^{T_\tau-1} \gamma^{\,s-t} c_s + \gamma^{\,T_\tau-t}\, c_{\mathrm{term}}(\rho(\tau)).
\end{equation}

For failures, this target includes accumulated cost and the terminal penalty rather than treating truncation length as remaining time. The value is $V=-C$, and downstream credit is computed from the corrected cost.

\subsection{GLC-Critic: Global-Local Cooperative Value-Correction Critic}

GLC-Critic combines local temporal evidence with a phase--progress coordinate to correct the expected remaining cost of a single-frame distributional base critic.

\begin{figure}[!t]
\centering
\includegraphics[width=0.90\columnwidth]{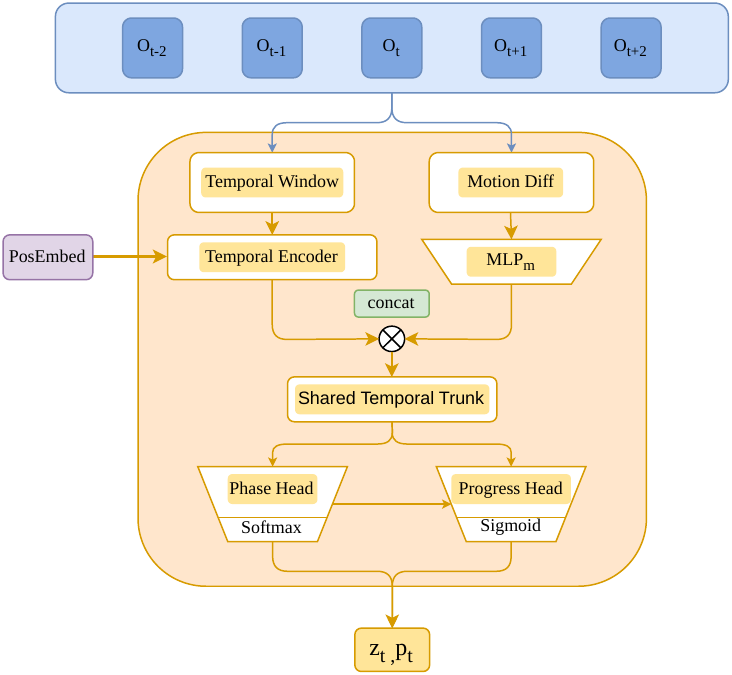}
\caption{Temporal and phase--progress branches of the GLC-Critic. Local observations and motion differences produce the coordinates used for value correction.}
\label{fig:critic_temporal}
\end{figure}

\textbf{Multi-task organization.} The visual and temporal backbones are shared across $K$ tasks, while phase, progress, and fusion heads are task-specific. Credit thresholds are also computed per task to avoid cross-task selection bias.

\textbf{Local temporal evidence encoding.} A trainable visual encoder extracts $f_t=E_v(o_t)$ from the image, proprioception, and instruction. A Transformer with positional embeddings \citep{vaswani2017attention} aggregates a centered local window:

\begin{equation}
\label{eq:temporal_evidence}
\begin{aligned}
W_t &= [f_{t-r},\ldots,f_{t+r}],\\
h_t^{\mathrm{temp}} &= \mathrm{TempEnc}(W_t).
\end{aligned}
\end{equation}

An additional branch encodes adjacent-frame changes:

\begin{equation}
\label{eq:motion_evidence}
h_t^{\mathrm{motion}} = \mathrm{MLP}_m([f_t;\, f_t - f_{t-1};\, f_{t+1} - f_t]).
\end{equation}

Both branches are shared across tasks. Their centered context $\mathcal{H}_{\mathrm{off}}$ is used only by the non-causal offline labeler; future observations never enter the deployed policy.

\textbf{Phase and progress prediction.} With $h_t=\mathrm{MLP}_p([h_t^{\mathrm{temp}};h_t^{\mathrm{motion}}])$, task-specific heads predict phase $z_t$ and intra-phase progress $p_t$:

\begin{equation}
\label{eq:phase_progress}
\begin{aligned}
z_t &= \mathrm{softmax}\big(W_z^{(k)}h_t\big) \in \mathbb{R}^{K_{\mathrm{ph}}},\\
p_t &= \sigma\!\left(W_p^{(k)}[h_t;E^{(k)\top}z_t]\right) \in [0,1],
\end{aligned}
\end{equation}
where $E^{(k)}$ embeds the phases of task $k$ and $K_{\mathrm{ph}}\in[3,5]$. Motion differences affect value correction only through this temporal coordinate.

\textbf{Phase--progress supervision.} Simulation boundaries come from privileged semantic predicates and real-robot boundaries are manually annotated. Qwen3-VL estimates intra-phase progress only within the final incomplete phase of failed trajectories, and every estimate is manually reviewed \citep{bai2025qwen3vl,chen2026sarm}. Privileged states supervise offline heads but never enter critic or policy observations; the complete annotation protocol is provided in the supplementary material.

\textbf{Remaining-cost value correction.} Following distributional value modeling \citep{bellemare2017distributional}, we discretize remaining cost into $B$ shared bins $\{b_j\}$. The base critic outputs $\ell^{\mathrm{raw}}=F_\theta(f_t)\in\mathbb{R}^B$ and

\begin{equation}
\label{eq:raw_distribution}
P_{\mathrm{raw}}(y_t = b_j \mid f_t) = \mathrm{softmax}(\ell^{\mathrm{raw}})_j,
\end{equation}
with expectation $\hat C^{\mathrm{raw}}=\sum_j b_jP_{\mathrm{raw}}(y_t=b_j\mid f_t)$. We correct this expectation using the task-specific temporal coordinate:

\begin{equation}
\label{eq:scalar_correction}
\begin{aligned}
\hat{C}(\mathcal{H}_{\mathrm{off}})
  &= \hat{C}^{\mathrm{raw}} + \lambda\,\Delta_t,\\
\Delta_t
  &= G_\psi^{(k)}\big([\ell^{\mathrm{raw}};\,z_t;\,p_t]\big),
\end{aligned}
\end{equation}

Here $G_\psi^{(k)}$ is a lightweight MLP and $V=-\hat C$. A shared scalar logit shift is ineffective because $\mathrm{softmax}(\ell+\kappa\mathbf1)=\mathrm{softmax}(\ell)$. Non-identical bin-wise offsets instead require $B$ correction outputs and introduce a higher-dimensional fitting burden, whereas downstream credit uses only the expectation. We therefore retain the base distribution and directly learn its scalar expected-cost correction. The base and phase--progress heads are first trained with

\begin{equation}
\label{eq:critic_objective}
\mathcal{L}_{\mathrm{critic}} = \mathcal{L}_{\mathrm{cost}} + \alpha_z \mathcal{L}_{\mathrm{phase}} + \alpha_p \mathcal{L}_{\mathrm{progress}},
\end{equation}
where $\mathcal{L}_{\mathrm{cost}}$ and $\mathcal{L}_{\mathrm{phase}}$ are categorical cross-entropy losses, and $\mathcal{L}_{\mathrm{progress}}$ is a smooth-L1 loss. With phase and progress frozen, $G_\psi$ is then trained on the corrected expectation:

\begin{equation}
\label{eq:correction_objective}
\mathcal{L}_{\mathrm{corr}} = \big( \hat{C}_t - C_t \big)^2,
\end{equation}
where $C_t$ is the normalized remaining-cost target.

\textbf{Step-level credit score.} From the corrected cost, we define

\begin{equation}
\label{eq:credit_score}
s_t = \hat{C}_t - \Big[ \sum_{i=0}^{H-1} \gamma^{\,i} c_{t+i} + \gamma^{H}\, \hat{C}_{t+H} \Big].
\end{equation}

Positive $s_t$ means that the observed $H$-step behavior improves on the conditional expectation; negative $s_t$ indicates additional cost. It supports task-wise quantile ranking but is not claimed to be an unbiased policy-gradient advantage. State-independent value shifts change $s_t$ by at most a task-wise constant, leaving its quantile labels unchanged; the derivation is given in the supplementary material.

\subsection{PPD: Progressive Policy Distillation}

PPD routes each round between PR-CFG \citep{ho2022classifierfree,frans2025diffusionguidance} at an average rollout SR of approximately $50\%$ or below and FACD above this nominal transition boundary. The reversible decision is recomputed across tasks every round. Both modes share a flow-matching policy \citep{lipman2023flow} and differ only in credit-conditioned sample use.

\textbf{Unified action objective.} For observation $o_i$, action chunk $a_i$, condition $g_i$, $\epsilon\sim\mathcal N(0,I)$, and $\tau\sim\mathcal U(0,1)$, define

\begin{equation}
\label{eq:flow_interpolation}
a_i^\tau = (1-\tau)\epsilon + \tau a_i.
\end{equation}
and optimize

\begin{equation}
\label{eq:flow_matching}
\mathcal{L}_{\mathrm{FM}}(o_i, a_i, g_i) = \mathbb{E}_{\epsilon, \tau}\Big[ \big\| v_\theta(a_i^\tau, \tau \mid o_i, g_i) - (a_i - \epsilon) \big\|^2 \Big],
\end{equation}
where $g_i\in\{\varnothing,\mathrm{pos},\mathrm{neg}\}$.

\textbf{PR-CFG.} At limited rollout coverage, the task-wise upper quantile defines reliable positives

\begin{equation}
\label{eq:positive_set}
\mathcal{P}^{(k)} = \{ i \in \mathcal{D}^{(k)} \mid s_i \ge \tau_{\mathrm{pos}}^{(k)} \}.
\end{equation}
and the objective is

\begin{equation}
\label{eq:prcfg_objective}
\begin{aligned}
\mathcal{L}_{\mathrm{PR}} =& \mathbb{E}_{i \sim \mathcal{D}}\bigl[\mathcal{L}_{\mathrm{FM}}(o_i, a_i, \varnothing)\bigr] \\
&+ \beta_{\mathrm{pos}}\, \mathbb{E}_{k}\, \mathbb{E}_{i \sim \mathcal{P}^{(k)}} \\
&\qquad \bigl[\mathcal{L}_{\mathrm{FM}}(o_i, a_i, \mathrm{pos})\bigr],
\end{aligned}
\end{equation}

At inference, positive residual guidance is

\begin{equation}
\label{eq:prcfg_guidance}
v_{\mathrm{PR}} = v_{\mathrm{unc}} + \eta\,(v_{\mathrm{pos}} - v_{\mathrm{unc}}),
\end{equation}
which reinforces high-credit behavior without using negative credit.

\textbf{FACD.} Above $50\%$ coverage, samples receive positive or negative conditions using task-wise credit thresholds; conditions are dropped with probability $0.3$ to obtain $\tilde g_i$. The objective is

\begin{equation}
\label{eq:facd_objective}
\begin{aligned}
\mathcal{L}_{\mathrm{FACD}} =& \mathbb{E}_{i \sim \mathcal{D}}\Bigl[ \mathcal{L}_{\mathrm{FM}}(o_i, a_i, \varnothing) \\
&+ w \cdot \mathbf{1}\{\tilde{g}_i \neq \varnothing\} \\
&\qquad \cdot \mathcal{L}_{\mathrm{FM}}(o_i, a_i, \tilde{g}_i) \Bigr],
\end{aligned}
\end{equation}
with $w=1$. Inference combines attraction to positive and repulsion from negative credit:

\begin{equation}
\label{eq:facd_guidance}
\begin{aligned}
v_{\mathrm{FACD}} &= v_{\mathrm{unc}} + \eta_{+}(v_{\mathrm{pos}} - v_{\mathrm{unc}}) \\
&\quad - \eta_{-}(v_{\mathrm{neg}} - v_{\mathrm{unc}}),
\end{aligned}
\end{equation}
where $\eta_+$ and $\eta_-$ control the two directions.

\textbf{Positive-sample protection.} Credit filtering remains task-specific. Failed-episode candidates enter the positive set in descending-credit order but are capped at a fraction $\rho_{\mathrm{fail}}$; expert demonstrations are labeled positive whenever included. Exact thresholds and fallback rules are provided in the supplementary material.

\textbf{Iterative loop.} At round $R$, the critic uses the accumulated buffer $\mathcal D_R=\cup_{i\leq R}\Delta_i$, whereas policy distillation uses only newly added $\Delta_R$. The router selects the objective, the policy VLM backbone remains frozen, and the updated policy collects $\Delta_{R+1}$. Full round-wise details appear in the supplementary material.

\section{Experiments}

We evaluate PACE on 10 LIBERO-Long tasks \citep{liu2023libero} and three real-robot tasks. The critic uses SigLIP-SO400M \citep{zhai2023sigmoid} and Gemma-3-270M \citep{gemmateam2025gemma3}; the policy uses $\pi_{0.5}$-Base \citep{physicalintelligence2025pi05}.

\subsection{Experimental Setup}

\textbf{Environments and protocol.} LIBERO-Long contains 10 tasks with 3--5 semantic phases; observations combine third-person RGB, proprioception, and language, and actions control 7-DoF end-effector pose and the gripper. Real-robot evaluation uses an AgileX PiPER dual-arm platform on Table Wiping, Towel Folding, and Mug Placement and Knob Turning. Platform, task, and hardware details are provided in the supplementary material.

\begin{figure}[!t]
\centering
\includegraphics[width=\columnwidth]{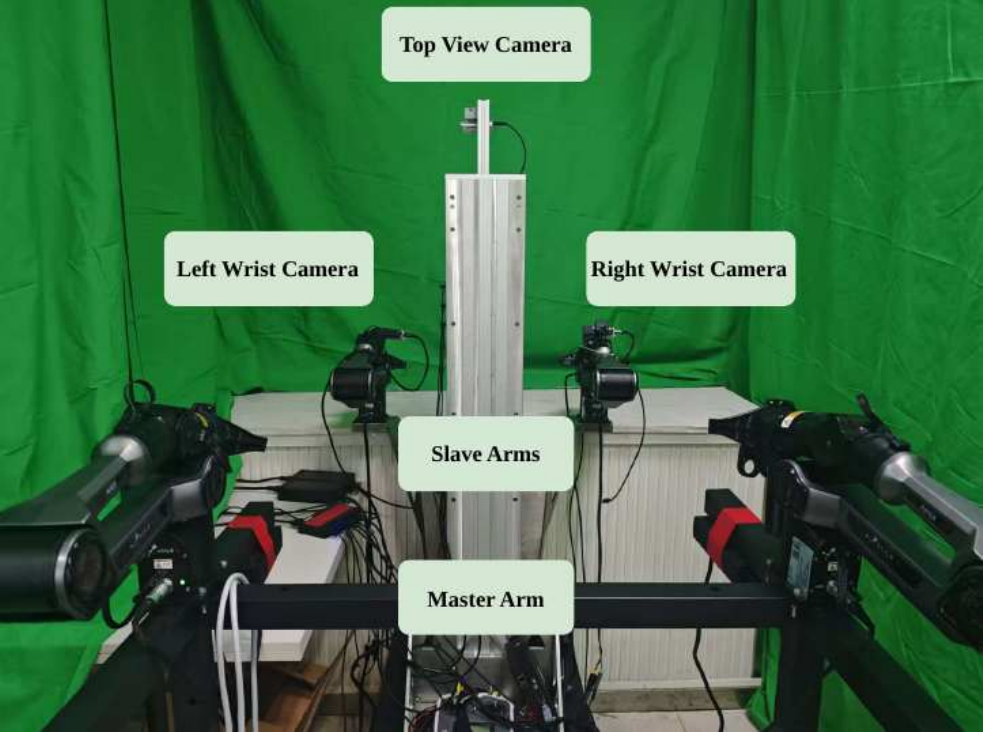}
\caption{Real-robot experimental platform with dual AgileX PiPER slave arms, bilateral master arms for teleoperation, and three camera views.}
\label{fig:real_robot_setup}
\end{figure}

In simulation, all methods start from 10 expert demonstrations per task. ReCAP and PACE share the same $\pi_{0.5}$ base policy, expert demonstrations, round-wise rollouts, and policy-update budget. Simulation uses four collection-and-update rounds: critic training accumulates valid data, whereas each policy update uses only the new round and applies the reversible $50\%$ router. Real-robot training uses 100 teleoperated demonstrations per task plus human-in-the-loop corrections. Full schedules appear in the supplementary material.

\textbf{Baselines and evaluation.} We compare SFT using OpenVLA \citep{kim2024openvla}, Xiaomi-Robotics-0 \citep{cai2026xiaomirobotics0}, or $\pi_{0.5}$; ReCAP \citep{amin2025pi06star}; and PACE. For PACE, the reported simulation results are computed over three independently trained policy checkpoints with distinct training seeds. Each checkpoint is evaluated with a fixed evaluation seed on the same ordered set of 20 initial states per task. All training and evaluation seeds are provided in the supplementary material. Each real-robot evaluation uses 50 trials per task. SR is trial success, and AvgT measures successful execution in simulation steps or real-robot seconds. Critic predictions are inverse-mapped before MAE; Held-out MAE pools validation frames, whereas Boundary MAE uses frames within $\pm10$ steps of semantic transitions. Failed returns additionally include the fixed terminal penalty.

\subsection{Main Simulation Results}

\begin{table}[!ht]
\centering
\footnotesize
\setlength{\tabcolsep}{1.5pt}
\caption{Main LIBERO-Long results. Critic errors use inverse-mapped return units; Boundary MAE uses $\pm10$-step transition neighborhoods.}
\label{tab:main_sim}
\begin{tabular}{@{}lcccc@{}}
\toprule
Method & \shortstack{Avg.\ SR\\(\%) $\uparrow$} & \shortstack{AvgT\\(steps) $\downarrow$} & \shortstack{Held-out\\MAE $\downarrow$} & \shortstack{Boundary\\MAE $\downarrow$} \\
\midrule
SFT (OpenVLA) & $52.7\pm1.6$ & $310.8\pm78.4$ & --- & --- \\
SFT (XR-0) & $57.2\pm1.3$ & $298.7\pm75.1$ & --- & --- \\
SFT ($\pi_{0.5}$) & $63.3\pm1.3$ & $276.9\pm68.3$ & --- & --- \\
ReCAP & $73.8\pm1.8$ & $254.26\pm63.54$ & 133.03 & 136.02 \\
\textbf{PACE} & \textbf{$83.3\pm2.5$} & \textbf{$252.55\pm69.07$} & \textbf{97.78} & \textbf{77.01} \\
\bottomrule
\end{tabular}
\end{table}

\begin{figure}[!t]
\centering
\begin{minipage}[t]{0.49\linewidth}
\centering
\includegraphics[width=\linewidth]{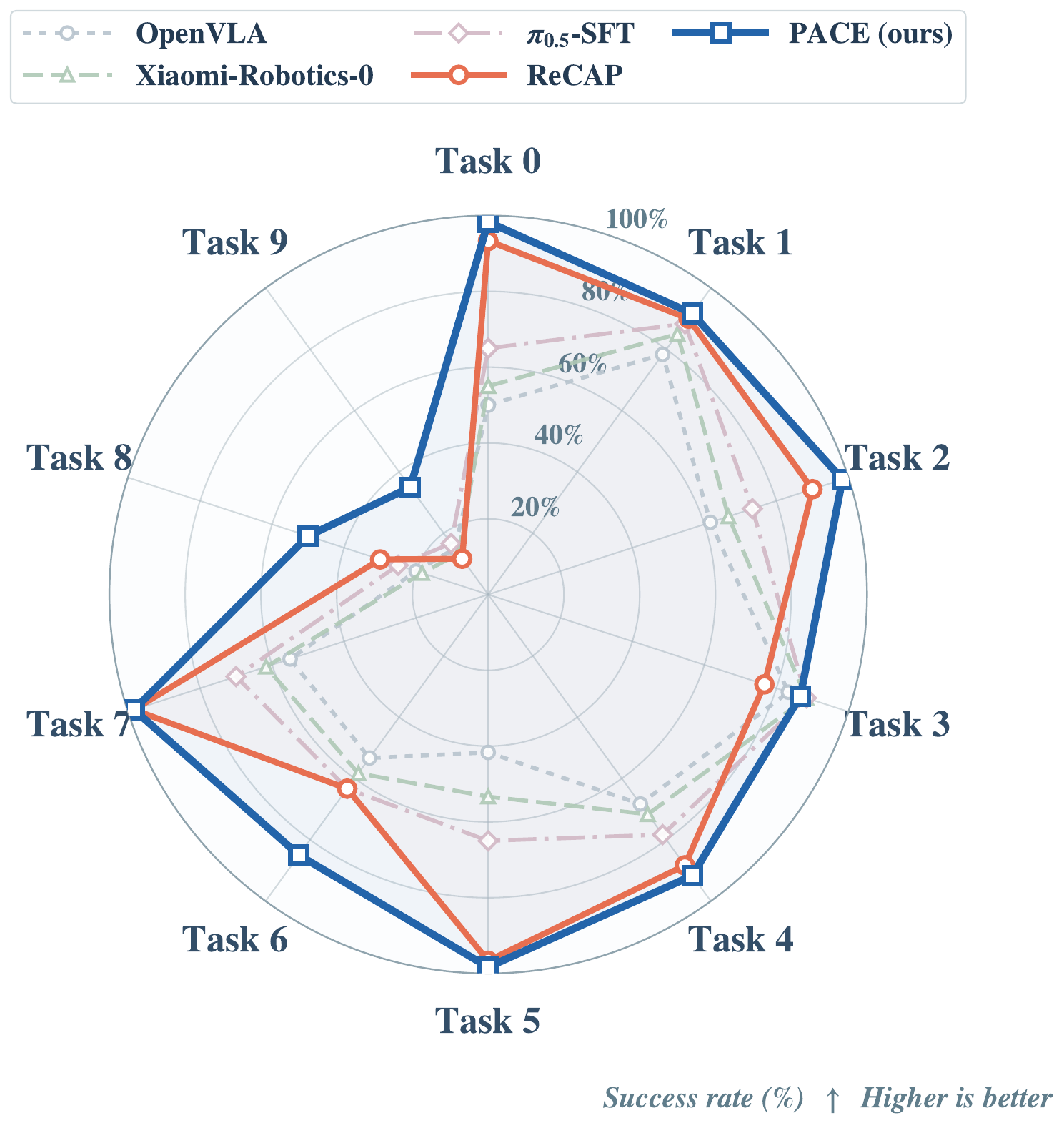}
\footnotesize (a) Task-wise success rate
\end{minipage}\hfill
\begin{minipage}[t]{0.49\linewidth}
\centering
\includegraphics[width=\linewidth]{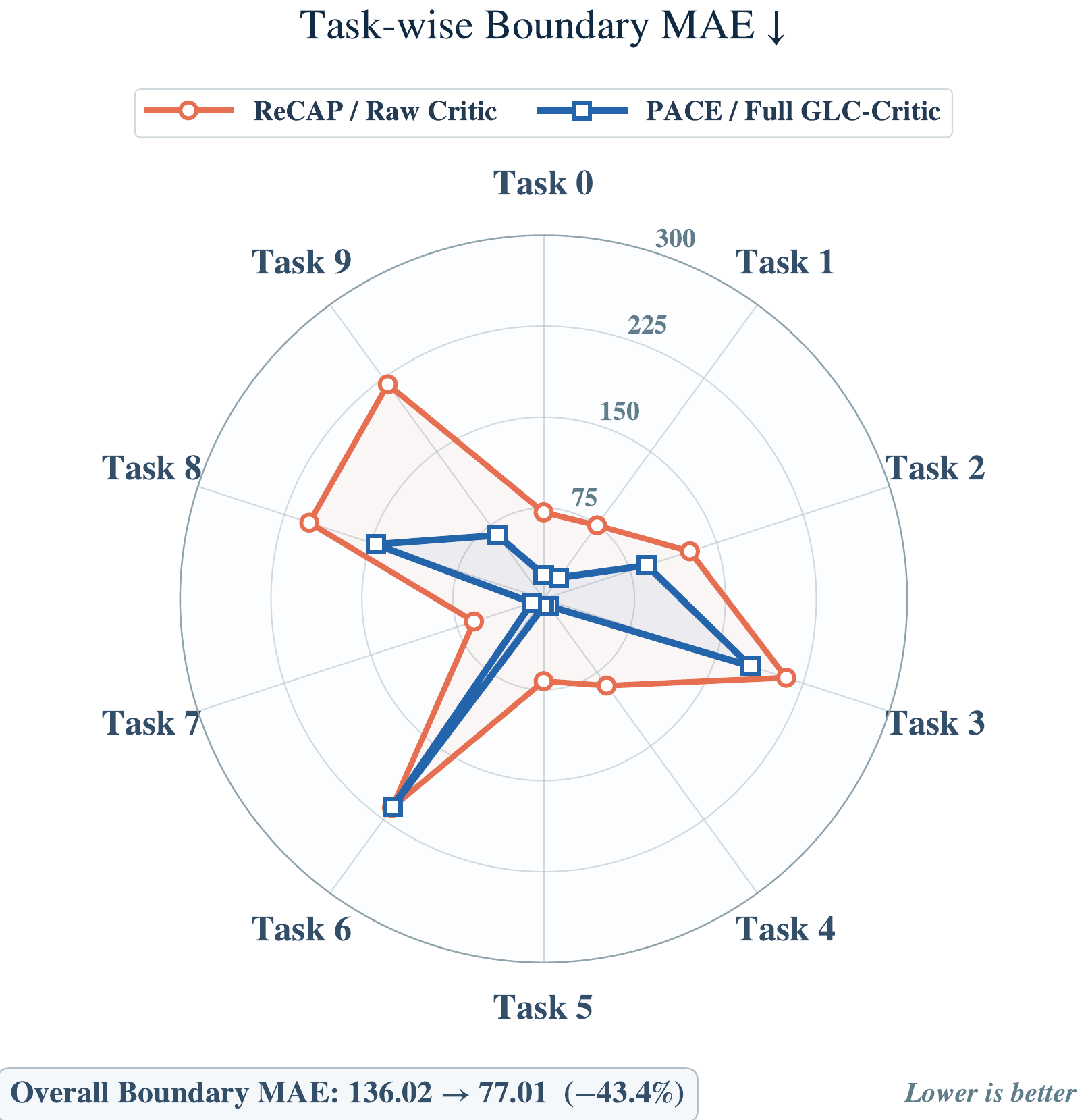}
\footnotesize (b) Task-wise Boundary MAE
\end{minipage}
\caption{Task-wise LIBERO-Long SR (a; higher is better) and Boundary MAE (b; lower is better).}
\label{fig:sim_radars}
\end{figure}

PACE reaches $83.3\pm2.5\%$ SR, improving over ReCAP by 9.5 points while maintaining comparable AvgT (Table~\ref{tab:main_sim}). It also exceeds all three SFT variants by at least 20 points, showing that the gain is not explained solely by the choice of pretrained VLA backbone. The nearly unchanged AvgT relative to ReCAP indicates that PACE primarily turns previously incomplete rollouts into successful ones rather than obtaining its SR gain by prolonging already successful executions. Figure~\ref{fig:sim_radars}(a) further shows that the gain is distributed across the 10 tasks rather than driven only by the average.

Using only the independently measured first evaluation of each method, we conduct a paired two-sided Wilcoxon signed-rank test over the task-wise SRs of the 10 LIBERO-Long tasks. Following the Wilcoxon convention, two zero-difference pairs are excluded, tied absolute differences are assigned midranks, and the exact $p$-value is computed by enumerating all $2^8$ sign assignments of the remaining nonzero signed ranks. PACE improves over ReCAP on 8 tasks, ties on 2 tasks, and degrades on none. The improvement is statistically significant ($W=0$, exact $p=0.0078$), with a mean paired gain of 9.5 percentage points and a median paired gain of 5.0 percentage points.

On 60 validation trajectories comprising 17,671 frames, GLC-Critic reduces Held-out MAE from 133.03 to 97.78. Near privileged semantic transitions, Boundary MAE falls from 136.02 to 77.01 (43.4\%), boundary return-ranking correlation rises from 0.745 to 0.871, and task-macro Boundary MAE improves on all 10 tasks (Fig.~\ref{fig:sim_radars}(b)). The relative reduction is larger at boundaries than over all frames, matching the intended role of phase--progress correction: disambiguating visually similar states that occur before and after a task transition. Together with the SR improvement obtained from the corrected critic, these results connect more accurate remaining-cost estimates to the downstream policy gain without claiming that value accuracy alone determines control performance.

\FloatBarrier
\subsection{Critic Architecture Ablation}

We remove each GLC-Critic component under the same data and policy-update protocol. All correction variants improve substantially over the single-frame Raw Critic (Table~\ref{tab:critic_ablation}). Removing temporal aggregation or motion differences raises both MAEs and lowers SR, confirming that a single observation is insufficient to recover the local direction of task evolution. Removing either coordinate head produces a smaller but consistent degradation. In particular, the progress-head ablation remains the strongest reduced variant, whereas the complete model still gives the lowest overall and boundary errors. The full model therefore benefits from complementary temporal, motion, phase, and intra-phase signals rather than depending on one dominant branch.

\begin{table}[!t]
\centering
\footnotesize
\setlength{\tabcolsep}{2pt}
\caption{GLC-Critic ablation under the protocol of Table~\ref{tab:main_sim}.}
\label{tab:critic_ablation}
\begin{tabular}{@{}p{0.43\columnwidth}ccc@{}}
\toprule
Critic Variant & \shortstack{SR (\%)\\$\uparrow$} & \shortstack{Held-out\\MAE $\downarrow$} & \shortstack{Boundary\\MAE $\downarrow$} \\
\midrule
Raw Critic & $73.8\pm1.8$ & 133.03 & 136.02 \\
w/o temporal aggregation & $81.6\pm2.3$ & 99.31 & 78.03 \\
w/o motion differences & $80.9\pm2.2$ & 100.01 & 79.43 \\
w/o phase head $z$ & $81.2\pm2.2$ & 100.17 & 78.73 \\
w/o progress head $p$ & $82.7\pm2.4$ & 98.03 & 77.23 \\
\textbf{Full GLC-Critic} & \textbf{$83.3\pm2.5$} & \textbf{97.78} & \textbf{77.01} \\
\bottomrule
\end{tabular}
\end{table}

\subsection{Progressive Distillation Curriculum Ablation}

Each comparison starts from the same incoming checkpoint and uses identical $\Delta_R$, critic predictions, and update budget. PR-CFG is stronger at low initial coverage: it leads FACD by 5.5 points in the first simulation round and by 10.7 points in the first real-robot round. Once the incoming policy covers more successful behavior, FACD becomes stronger in simulation round~2 and real-robot rounds~3--4 (Table~\ref{tab:curriculum_ablation}).

The largest late-round gap appears in real-robot round~4, where using both credit directions reaches $81.8\%$ versus $67.8\%$ with positive-only guidance. This reversal motivates the router: early updates protect useful pretrained behavior, while later updates exploit negative credit after rollout coverage improves.

\begin{table}[!ht]
\centering
\footnotesize
\setlength{\tabcolsep}{3pt}
\caption{Controlled PPD ablation with shared initialization, data, critic predictions, and update budget. $^{*}$ marks a significant paired two-sided Wilcoxon difference over task--evaluation blocks after Holm correction across the six rows.}
\label{tab:curriculum_ablation}
\begin{tabular}{@{}p{0.52\columnwidth}cc@{}}
\toprule
Update (shared init.) & \shortstack{PR-CFG\\SR (\%) $\uparrow$} & \shortstack{FACD\\SR (\%) $\uparrow$} \\
\midrule
\multicolumn{3}{@{}l}{\textit{Simulation}} \\
Round 1 on $\Delta_1$ (base, $44.2$) & \textbf{$54.5\pm2.0^{*}$} & $49.0\pm2.5$ \\
Round 2 on $\Delta_2$ (same R1 PR-CFG, $54.5$) & $61.0\pm2.0$ & \textbf{$64.5\pm1.5^{*}$} \\
\addlinespace[2pt]
\multicolumn{3}{@{}l}{\textit{Real Robot}} \\
Round 1 on $\Delta_1$ (base, $16.4$) & \textbf{$32.9\pm1.9^{*}$} & $22.2\pm2.3$ \\
Round 2 on $\Delta_2$ (same R1 PR-CFG, $32.9$) & \textbf{$49.6\pm2.1^{*}$} & $41.3\pm0.7$ \\
Round 3 on $\Delta_3$ (same R2 PR-CFG, $49.6$) & $60.4\pm0.4$ & \textbf{$62.0\pm1.3$} \\
Round 4 on $\Delta_4$ (same R3 FACD, $62.0$) & $67.8\pm0.8$ & \textbf{$81.8\pm1.7^{*}$} \\
\bottomrule
\end{tabular}
\end{table}

At 49.6\% incoming SR, PR-CFG and FACD perform comparably (60.4\% vs. 62.0\%), placing this round at the nominal 50\% transition boundary; we therefore use FACD thereafter.

\FloatBarrier

\subsection{Real-Robot Results}

\begin{figure}[!t]
\centering
\includegraphics[width=\linewidth]{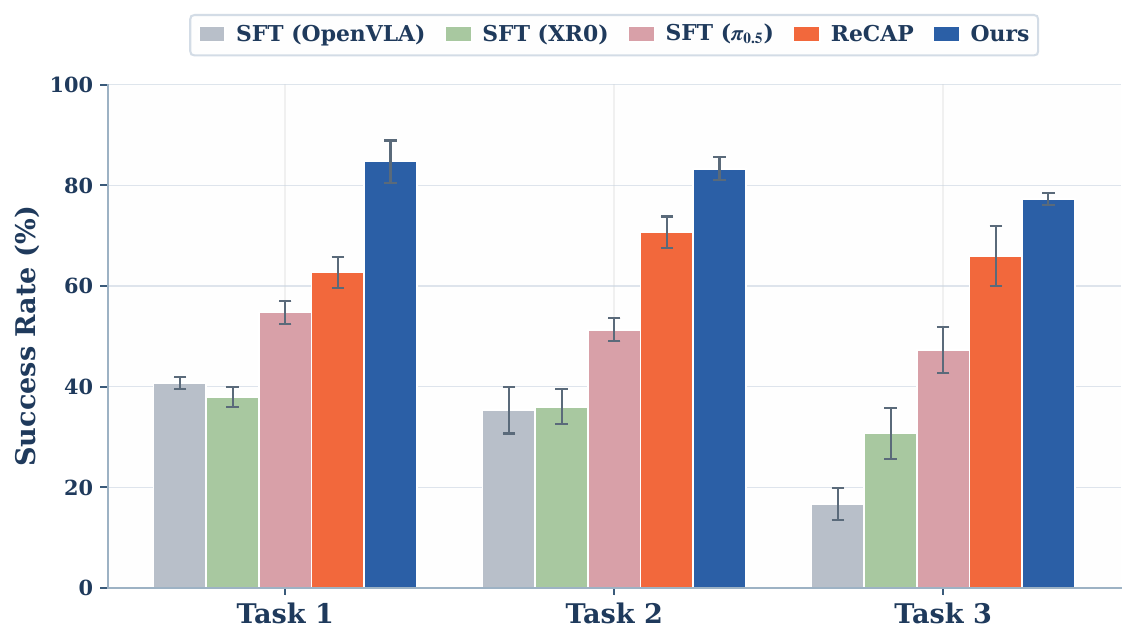}
\caption{Real-robot task-wise SR over three repeated evaluations; higher is better.}
\label{fig:real_robot_sr}
\end{figure}

PACE reaches 81.8\% task-macro SR, 15.3 points above ReCAP, and reduces task-average successful-trial AvgT from 76.3 to 58.7 seconds (Table~\ref{tab:real_robot}). It achieves the highest SR and lowest AvgT on all three tasks (Fig.~\ref{fig:real_robot_sr}), including the longer Mug Placement and Knob Turning sequence.

Across the nine task--evaluation blocks, all SR improvements over the four baselines are significant under two-sided exact sign-flip tests with Holm correction (adjusted $p=0.0156$, the minimum attainable with nine blocks). Per-task two-sided Fisher exact tests on 150 pooled trials per method corroborate these gains (Holm-adjusted $p\leq0.04$).

The critic improvement is concentrated near task transitions: Held-out MAE decreases from 68.17 to 18.89, while Boundary MAE decreases from 58.73 to 4.12. This pattern agrees with simulation and suggests that phase-aware correction transfers to contact-rich physical execution despite sensing and actuation noise.

\begin{table*}[t]
\centering
\footnotesize
\caption{Task-wise real-robot results. AvgT is successful-trial duration in seconds. Tasks 1--3 are Table Wiping, Towel Folding, and Mug Placement and Knob Turning. For SR, $^{*}$ indicates that PACE significantly outperforms the corresponding baseline on that task (two-sided Fisher exact test on 150 pooled trials per method, Holm-adjusted $p\leq0.04$).}
\label{tab:real_robot}
\setlength{\tabcolsep}{3pt}
\begin{tabular}{@{}lcccccccc@{}}
\toprule
Method & \multicolumn{2}{c}{Task 1} & \multicolumn{2}{c}{Task 2} & \multicolumn{2}{c}{Task 3} & \shortstack{Held-out\\MAE $\downarrow$} & \shortstack{Boundary\\MAE $\downarrow$} \\
\cmidrule(lr){2-3}\cmidrule(lr){4-5}\cmidrule(lr){6-7}
& SR (\%) $\uparrow$ & AvgT (s) $\downarrow$ & SR (\%) $\uparrow$ & AvgT (s) $\downarrow$ & SR (\%) $\uparrow$ & AvgT (s) $\downarrow$ & & \\
\midrule
SFT (OpenVLA) & $40.7\pm1.2^{*}$ & $89.7\pm1.7$ & $35.3\pm4.6^{*}$ & $84.8\pm3.2$ & $16.7\pm3.1^{*}$ & $90.4\pm3.5$ & --- & --- \\
SFT (XR-0) & $38.0\pm2.0^{*}$ & $83.8\pm10.9$ & $36.0\pm3.5^{*}$ & $82.9\pm6.6$ & $30.7\pm5.0^{*}$ & $77.1\pm6.1$ & --- & --- \\
SFT ($\pi_{0.5}$) & $54.7\pm2.3^{*}$ & $78.2\pm3.6$ & $51.3\pm2.3^{*}$ & $76.7\pm3.1$ & $47.3\pm4.6^{*}$ & $74.9\pm4.5$ & --- & --- \\
ReCAP & $62.7\pm3.1^{*}$ & $72.5\pm4.3$ & $70.7\pm3.1^{*}$ & $76.9\pm3.9$ & $66.0\pm6.0^{*}$ & $79.5\pm3.3$ & 68.17 & 58.73 \\
\textbf{PACE (ours)} & \textbf{$84.7\pm4.2$} & \textbf{$53.0\pm9.8$} & \textbf{$83.3\pm2.3$} & \textbf{$58.9\pm16.6$} & \textbf{$77.3\pm1.2$} & \textbf{$64.3\pm11.5$} & \textbf{18.89} & \textbf{4.12} \\
\bottomrule
\end{tabular}
\end{table*}

\FloatBarrier

\FloatBarrier
\subsection{Critic Behavior and Failure Analysis}

Figure~\ref{fig:credit_motivation} provides a trajectory-level view of the aggregate critic improvements. On the representative successful rollout, the Raw Critic remains systematically below the target remaining return and responds weakly to completed phases. The fused estimate instead tracks the long-term increase toward successful termination and adjusts more clearly around the annotated transitions, reducing episode MAE from 205.77 to 43.22 and the magnitude of the signed bias from 205.77 to 30.19. This behavior is consistent with the larger validation-set improvement at phase boundaries: local motion identifies whether a visually similar state is approaching or leaving a milestone, while the phase--progress coordinates provide a task-level reference for the scalar correction.

The incomplete rollout illustrates a harder regime rather than a uniformly solved case. Here the target includes the terminal failure penalty, and repeated or regressive behavior can revisit similar observations without making equivalent task progress. The Fused Critic still reduces episode MAE from 379.32 to 214.62, but its predictions remain variable and retain a positive bias. In particular, uncertainty in local progress can produce transient over-correction when the trajectory stalls near a transition or moves back toward an earlier configuration. These residual errors explain why the correction is applied to the expected cost rather than treated as a replacement for the distributional base critic.

Completed milestones provide stable phase anchors, whereas unresolved failures leave progress to local dynamics alone; this asymmetry explains the residual errors on difficult failed rollouts.

For downstream distillation, absolute calibration is not the only relevant property: credit labels depend on the ordering induced by temporal cost differences. Task-wise quantile selection makes this labeling invariant to a state-independent value shift, but it cannot remove state-dependent ranking errors. PACE therefore limits failed-episode positives, retains demonstration positives when available, and delays full positive--negative conditioning until rollout coverage is sufficient. The critic and curriculum play complementary roles: the former improves discrimination near semantic transitions, whereas the latter limits the influence of the remaining critic errors on policy updates.

These observations also delimit the current scope. Phase supervision is task-specific, and failed trajectories require manual review when the final incomplete phase cannot be determined from verified boundaries. Although privileged predicates are used only to construct offline auxiliary targets and never enter critic or policy observations, extending PACE to less structured tasks will require reliable phase discovery without such supervision. Representative real-robot phase and critic trajectories, together with the complete annotation protocol, are provided in the supplementary material.

\section{Conclusion}

\looseness=-1
PACE is a credit-assignment framework for post-training VLA models on long-horizon manipulation. It targets unreliable step-level credit under sparse terminal rewards by improving both credit estimation and credit utilization. The GLC-Critic combines a discretized distributional base critic with phase--progress-aware scalar correction of its expected remaining cost, yielding step-level credit ranked by task-wise quantiles. Progressive Policy Distillation (PPD) converts credit into positive and negative conditions and selects their use through a coverage-adaptive, reversible curriculum. On LIBERO-Long, PACE reduces held-out per-frame remaining-cost MAE by 26.5\% and reduces the error near semantic phase transitions by 43.4\%. The average success rate improves from 73.8\% to 83.3\% over the best-performing compared baseline. Results on real dual-arm tasks follow the same trend. Limitations include task-specific phase annotation and residual over-correction on a few hard tasks. Future work will explore automated phase discovery across more tasks and embodiments.

\bibliography{aaai2027}

\end{document}